%% file: acl_latex.tex
\documentclass[11pt]{article}

\usepackage[preprint]{acl}

\usepackage{times}
\usepackage{latexsym}

\usepackage[T1]{fontenc}

\usepackage[utf8]{inputenc}

\usepackage{microtype}

\usepackage{inconsolata}

\usepackage{graphicx}
\usepackage{float}
\usepackage{fancyvrb}
\usepackage{graphicx}
\usepackage{boldline} 
\usepackage{booktabs}
\usepackage{multirow}
\usepackage{sidecap}
\usepackage{subcaption}
\usepackage{amsmath}
\usepackage{amssymb}
\usepackage{amsfonts}
\usepackage{xcolor}
\usepackage{bm}
\usepackage{hyperref}
\usepackage{rotating}
\usepackage{adjustbox}
\usepackage{algorithm}
\usepackage{algpseudocode} 
\usepackage{array}
\definecolor{green}{rgb}{0.0, 0.5, 0.0}
\definecolor{red}{rgb}{0.82, 0.1, 0.26}
\definecolor{codegray}{gray}{0.9}

\usepackage{listings}
\definecolor{codegreen}{rgb}{0,0.6,0}
\definecolor{codegray}{rgb}{0.5,0.5,0.5}
\definecolor{codepurple}{rgb}{0.58,0,0.82}
\definecolor{backcolour}{rgb}{0.95,0.95,0.92}
\lstdefinestyle{mystyle}{
    backgroundcolor=\color{backcolour},   
    commentstyle=\color{codegreen},
    keywordstyle=\color{magenta},
    numberstyle=\tiny\color{codegray},
    stringstyle=\color{codepurple},
    basicstyle=\ttfamily\footnotesize,
    breakatwhitespace=false,         
    breaklines=true,                 
    captionpos=b,                    
    keepspaces=true,                 
    numbers=left,                    
    numbersep=5pt,                  
    showspaces=false,                
    showstringspaces=false,
    showtabs=false,                  
    tabsize=2
}
\usepackage{pifont}

\newlength{\Oldarrayrulewidth}

\definecolor{darkpurple}{RGB}{75, 0, 130}
\definecolor{darkyellow}{RGB}{230, 184, 0}

\usepackage[most]{tcolorbox}
\tcbuselibrary{breakable}   
\newtcolorbox{myshadowbox}{
    enhanced,
    colback=white, 
    colframe=black, 
    shadow={1mm}{-1mm}{0mm}{black!50!white}, 
    boxrule=0.5pt 
}

\newcounter{quotecount}

\usepackage{colortbl} 
\usepackage{tabularx} 

\newcounter{prop}

\newif\ifcoloredtext
\coloredtextfalse 

\usepackage{amsthm}
\newtheorem*{assumption*}{Assumption}

\usepackage{tikz}
\usetikzlibrary{arrows.meta, positioning, shapes.multipart, calc}
\usepackage{pgfplots}
\pgfplotsset{compat=1.18}

\title{When Does Memory Help Multi-Trajectory Inference for Tool-Use LLM Agents?}

\author{Xinzhe Li \\
  RMIT University \\
  \texttt{xinzhe.li@rmit.edu.au} \\\And
  Yaguang Tao \\
  RMIT University \\
  \texttt{yaguang.tao@rmit.edu.au} \\}

\begin{document}
\maketitle
\begin{abstract}
\input{sections/0.abstract}
\end{abstract}

\input{sections/1.intro.tex}

\input{sections/2.related-work.tex}
\input{sections/3.method.tex}
\input{sections/4.exp_setup.tex}

\input{sections/5.results-and-analysis}

\input{sections/6.discussion}
\input{sections/7.conclusion}

\section*{Limitations}

Our study evaluates a single policy model per benchmark (Haiku~3.5
for SQL tasks, Sonnet~4.6 for KGQA); the diversity dynamics that
drive the Raw Sibling and Reflection effects may differ for models
with different sampling behaviour at fixed temperature.
Sample sizes per cell range from 49 to 89 examples, which is
sufficient to detect effects of $\gtrsim 6$pp under McNemar's exact
test but not smaller ones.
We evaluate cross-sibling and cross-trajectory memory scopes but
not cross-task memory (knowledge transfer across different
questions), which may exhibit different dynamics.
LiTS-Fact uses non-selective retrieval (all extracted facts are
injected); a structural argument for why selective retrieval cannot
resolve the diversity--efficiency tradeoff is given in
Appendix~\ref{sec:appendix-retrieval}.

\section*{Ethics Statement}

This work involves no human subjects, no personal data, and no
deployment of autonomous agents in real-world settings.
All experiments use commercial LLM APIs (Anthropic Claude) on
publicly available benchmarks.
The total API cost is approximately \$1,384
(Appendix~\ref{app:compute-cost}).
We do not foresee direct negative societal impacts from the
findings; the memory methods studied are general-purpose
augmentations to LLM inference and do not introduce new
capabilities beyond what the underlying models already possess.
We used AI coding assistants (Claude) for code development and
paper drafting assistance.

\section*{Acknowledgements}

This research was supported by compute resources provided by the
RMIT Advanced Computing Ecosystem (RACE) through AWS cloud credits.

\bibliography{custom}

\appendix

\input{sections/app-rl-connection}

\input{sections/app-expansion}

\input{sections/app-method-motivation}

\input{sections/app-omitted-cells}

\input{sections/app-backprop}

\input{sections/app-search-settings}

\input{sections/app-methods-detail}

\input{sections/app-pipeline}

\input{sections/app-kgqa-subset}

\input{sections/app-mcnemar-pvalues}

\input{sections/app-pattern1-mechanism}

\input{sections/app-beam-giveup}

\input{sections/app-kgqa-fact-regression}

\input{sections/app-conflict}

\input{sections/app-retrieval}

\input{sections/app-efficiency}

\input{sections/app-compute-cost}

\end{document}

%% file: sections/0.abstract.tex
Multi-trajectory inference for tool-use LLM agents --- generating
multiple reasoning attempts and selecting among them --- benefits
from transferring knowledge across attempts so that later ones avoid
the pitfalls of earlier ones.
Existing cross-trajectory memory methods (trajectory-level
reflection, atomic fact extraction, raw observation injection) are
each evaluated under a single inference strategy on a single task,
making it unclear whether reported gains reflect properties of the
memory abstraction or of the inference method.
We propose a unified framework that decomposes memory along two
axes --- the \emph{scope} of transfer (within an expansion vs.\
across trajectories) and the \emph{abstraction} of the transferred
content --- and evaluate four methods under three inference
strategies (best-of-$N$, beam search, MCTS) on four tool-use
benchmarks spanning SQL, knowledge-graph, and CLI environments, in
a verifier-free setting that matches the deployment regime of
practical agents.
The experiment matrix identifies the inference method as a confound:
the same memory method produces statistically distinct results under
different inference strategies on the same examples.
Reflection reaches significance only under MCTS (not under
best-of-$N$); within-expansion injection (conditioning each
candidate on prior siblings' outcomes) helps only
diversity-starved beam search; and atomic fact extraction is
accuracy-neutral but shortens trajectories by 19--26\% on tasks
with reusable environmental structure.

%% file: sections/1.intro.tex
\section{Introduction}
\label{sec:intro}

Multi-trajectory inference --- generating many candidate reasoning
paths and selecting among
them~\citep{yao2023tree,hao-etal-2023-reasoning,gui2024bonbon} ---
consistently improves accuracy over single-trajectory agents on
tool-use benchmarks.
One way to further improve multi-trajectory inference is to
explicitly \emph{transfer information} across attempts (e.g.,
avoiding repeated mistakes, reusing discovered environmental
knowledge, or diversifying exploration), so that later attempts
are informed by earlier ones rather than sampling independently.

Memory augmentation provides this information channel. Existing
methods, however, span a heterogeneous design space:
Reflexion-style trajectory-level summaries of failed
attempts~\citep{shinn2023reflexion},
LATS-style reflections injected during tree
search~\citep{zhou2024language},
and Re-TRAC's structured state combining evidence, plans, and
failed-attempt records~\citep{zhu2026re}.
These methods vary in the \emph{abstraction} at which transferred
content is represented (strategy-level reflection vs.\ structured
facts), but all operate at the same \emph{scope}: transferring
knowledge across complete trajectories.
We identify scope as a second axis --- within-expansion transfer
(conditioning each sibling candidate on prior siblings' outcomes)
--- and show that the combination of scope and abstraction interacts
nontrivially with the inference strategy (best-of-$N$, beam search,
MCTS) and with task properties (serializable vs.\ non-serializable
environment, presence or absence of an inline verifier).

We focus on \emph{tool-use} agents: multi-step interactive reasoning
where the agent issues structured calls (e.g., SQL queries, shell
commands) to an external environment.
This setting supports both memory abstractions (observational
feedback for fact extraction; error signals for reflection) and
exposes a structural constraint: when the environment is
non-serializable~\citep{zainullina2025guided} (state cannot be
forked), beam search and MCTS are infeasible, leaving
memory-augmented best-of-$N$ as the only viable
multi-trajectory strategy.

We propose a unified framework that decomposes memory along the
scope$\times$abstraction axes, derives four memory methods as concrete
instantiations, and evaluates them in an experiment matrix
($4$ memory methods $\times$ $3$ inference strategies $\times$
$4$ benchmarks spanning $3$ tool-use environments, minus structurally inadmissible combinations).
Our setting is \emph{verifier-free}: the task verifier is offline,
applied at evaluation time only, matching the deployment regime of
practical tool-use agents. This distinguishes our analysis from prior memory-augmented
multi-trajectory work that relies on inline binary verifiers
(exact match, unit-test
pass)~\citep{shinn2023reflexion,zhou2024language}.

Our contributions:
\begin{itemize}
\item A unified scope$\times$abstraction framework for memory in
  tool-use tree search, from which existing methods and a new one
  (Raw Sibling) are derived as instantiations
  (\S\ref{sec:framework}).
\item A systematic empirical study under a verifier-free,
  deployment-faithful setup, covering memory $\times$ inference
  $\times$ task cells (\S\ref{sec:setup}).
\item Three findings on when memory helps:
  (F1) memory's accuracy effect is search-method-dependent ---
  Reflection reaches significance only under MCTS, while
  cross-sibling injection helps only diversity-starved beam search;
  (F2) under MCTS on the harder benchmark (KGQA), Reflection and
  Raw Sibling produce statistically indistinguishable accuracy
  despite operating on different memory abstractions;
  (F3) fact extraction is accuracy-neutral but improves
  \emph{efficiency} on tasks with reusable environmental structure
  (\S\ref{sec:results}).
\end{itemize}

%% file: sections/2.related-work.tex
\section{Background and Related Work}
\label{sec:related}

\paragraph{LLM tree search and multi-trajectory inference.}
Recent work applies search algorithms to LLM inference,
generating multiple reasoning trajectories and selecting the best.
Tree-of-Thoughts \cite{yao2023tree} uses BFS/DFS with LLM-generated evaluations.
RAP \cite{hao-etal-2023-reasoning} formulates reasoning as MCTS with an LLM world model.
ReST-MCTS* \cite{zhang2024restmcts} combines MCTS with process reward models
for step-level guidance.
Best-of-$N$ sampling \cite{gui2024bonbon} generates best-of-$N$
and selects via a verifier or reward model.
\citet{zainullina2025guided} formalize non-serializable environments
where state cannot be forked, showing that MCTS is infeasible
in such settings and proposing trajectory selection as an alternative.

\paragraph{Reflection and verbal memory in LLM agents.}
Reflexion \cite{shinn2023reflexion} stores verbal reflections from failed trials
in episodic memory to improve subsequent attempts.
LATS \cite{zhou2024language} integrates reflection into MCTS,
generating self-reflections after failed rollouts.
Re-TRAC \cite{zhu2026re} compresses each trajectory into a structured
state (evidence, uncertainties, plans) and conditions subsequent attempts on it ---
a unified representation combining our reflection and fact extraction abstractions.
RoT \cite{zhou2022reflection} extracts guidelines from prior search trees
to reduce repeated mistakes in future searches.
ExpeL \cite{zhao2024expel} accumulates cross-task insights
from success and failure trajectories.
R-MCTS \cite{yu2025exact} adds contrastive reflection to MCTS
for web navigation agents.
MC-DML \cite{shi2025monte} uses MCTS with a dynamic memory library
that stores reflections as PUCT priors for text game agents.

\paragraph{Atomic fact extraction.}
Fact-based memory has been explored in conversational and planning
settings.
mem0~\citep{chhikara2025mem0} extracts and consolidates atomic facts
from multi-session dialogues using vector-indexed storage and
similarity-based retrieval.
\citet{holt2025improving} apply atomic fact augmentation to lookahead search
in environment-grounded planning (ALFWorld, TextFrozenLake),
extracting task-critical facts from interaction trajectories.
Our LiTS-Fact adapts the mem0 extraction pipeline to the
multi-attempt tool-use search setting and provides the first
controlled comparison against trajectory-level reflection.

These works each propose a specific memory mechanism;
none systematically compares different memory abstractions
(raw observations vs.\ reflections vs.\ extracted facts)
within the same experimental setup.

\paragraph{Multi-agent memory sharing.}
Cross-agent memory collaboration has been explored via
shared memory pools \cite{gao2024memory},
contrastive trajectory distillation across heterogeneous models \cite{chang2026memcollab},
swarm-inspired role specialization \cite{li2025swarmsys},
and off-trajectory reasoning where injected partial traces
can act as distraction rather than guidance \cite{li2026offtrajectory}.
These approaches use model ensembles;
our work studies the same exploration-exploitation tradeoff
within a single agent across multiple attempts.

\paragraph{Connection to reinforcement learning.}
Our framework can be viewed as an inference-time analogue
of experience replay~\cite{lin1992self}:
processed experience (reflections, extracted facts)
is injected into the policy prompt rather than used
for gradient updates.
We discuss connections to in-context learning and
hindsight experience replay in
Appendix~\ref{sec:appendix-rl-connection}.

%% file: sections/3.method.tex
\section{A Unified Memory Framework}
\label{sec:framework}

We formalize context augmentation for LLM-based multi-trajectory reasoning.
Let $\pi_\theta(a \mid s)$ denote the base policy (an LLM generating candidate actions given state~$s$).
A \emph{context augmentor} modifies the policy by conditioning on additional context:
\begin{equation}
\label{eq:augmented-policy}
\pi_\theta(a \mid s, \mathcal{C}), \quad \text{where} \quad \mathcal{C} = \bigcup_{k=1}^{K} f_k\!\bigl(\mathcal{H}_k\bigr).
\end{equation}
Here $f_k$ is the $k$-th augmentor's \emph{analyze} function, and $\mathcal{H}_k$ is the history accessible to augmentor~$k$.
In practice, $\mathcal{C}$ is concatenated into the LLM prompt alongside~$s$.

Our framework decomposes memory along two orthogonal axes:
\emph{scope} (\S\ref{sec:scope}) --- what history is accessible to the augmentor,
and \emph{abstraction} (\S\ref{sec:abstraction}) --- how that history is transformed
before injection into the prompt.
Concrete memory methods (\S\ref{sec:methods}) are instantiations
of specific (scope, abstraction) pairs,
and multiple methods compose naturally via Eq.~\eqref{eq:augmented-policy}.

\subsection{Memory Scope}
\label{sec:scope}

The memory scope defines $\mathcal{H}_k$ --- what history the augmentor can access.
Let $\tau_i = (s_0, a_0, s_1, a_1, \ldots, s_T)$ denote the $i$-th trajectory
within a single search, and $\mathcal{T}_n$ the set of all trajectories
from the $n$-th task instance.

\paragraph{Cross-sibling (within expansion).}
During a single expansion of $N$ candidates at node~$s$,
the augmentor provides each candidate with the actions and observations
of previously sampled siblings:
\begin{equation}
\mathcal{H}_{\text{sib}}^{(i)} = \{(a_j, o_j)\}_{j < i}, \quad o_j = T(s, a_j),
\end{equation}
where $T$ is the transition function.
This requires \emph{interleaved expansion}: candidates are sampled sequentially,
each conditioned on the completed steps of prior siblings:
\begin{equation}
\label{eq:sibling-expand}
a_i \sim \pi_\theta\!\bigl(a \mid s,\; \mathcal{C},\; \{(a_j, o_j)\}_{j < i}\bigr).
\end{equation}
Cross-sibling scope is only applicable to inference methods
that expand multiple candidates at the same node
(beam search, MCTS), not to best-of-$N$.
Appendix~\ref{sec:appendix-expansion} contrasts this interleaved
expansion with the standard batch expansion used by methods without
cross-sibling scope.

\paragraph{Cross-trajectory (within search).}
The augmentor accesses trajectories from previous iterations of the same search:
\begin{equation}
\mathcal{H}_{\text{traj}} = \bigcup_{j < i} \tau_j.
\end{equation}
This includes both step-level fact sharing
(retrieving environmental discoveries from prior trajectories)
and trajectory-level profiling
(analyzing completed trajectories and injecting structured summaries).
Cross-trajectory scope applies to all three inference methods:
in best-of-$N$, each attempt is a ``trajectory''
whose knowledge transfers to subsequent attempts.

A third scope, \emph{cross-task} (across task instances),
is the native operating regime of cross-task memory systems
such as ExpeL~\cite{zhao2024expel} and mem0~\cite{chhikara2025mem0}.
We focus this paper on cross-trajectory memory within a single task instance
and leave a systematic study of cross-task memory to future work.

\subsection{Abstraction Level}
\label{sec:abstraction}

The abstraction level determines how history $\mathcal{H}_k$ is transformed
into context $\mathcal{C}_k = f_k(\mathcal{H}_k)$ before prompt injection.
We identify three levels, forming a spectrum from zero processing cost
to maximum compression:

\paragraph{Raw (no abstraction).}
The history is injected verbatim --- complete action--observation pairs
without any LLM summarization.
Cost: zero additional LLM calls.
Limitation: consumes context window proportional to history size;
only practical for small histories (e.g., a few sibling steps).

\paragraph{Reflection (trajectory-level summary).}
An LLM generates a natural-language summary of an entire trajectory,
capturing strategy-level lessons
(e.g., ``the C compilation approach failed due to missing libraries;
try a Python-based solution instead'').
Cost: one LLM call per trajectory.

\paragraph{Atomic fact extraction (step-level or batch).}
An LLM extracts discrete factual statements from observations
(e.g., ``config file located at \texttt{/etc/app/config.yaml}'',
``\texttt{apt-get install gcc} failed: package not found'').
Facts are deduplicated against existing entries via embedding-based
similarity at write time;
at inference, all facts collected from prior trajectories
are injected into the policy prompt
(we discuss alternative retrieval policies in \S\ref{sec:discussion}).
Cost: one LLM call per step (incremental) or per trajectory (batch),
plus embedding computation for deduplication.

\paragraph{Reflection vs.\ fact extraction.}
In best-of-$N$ with full-trajectory processing,
reflection and fact extraction share the same pipeline structure
(one LLM call on the complete trajectory, results injected into subsequent attempts)
but differ in \emph{what} the LLM is asked to produce:
reflection yields strategy-level advice,
while fact extraction yields observation-level atomic statements.
This makes them a controlled comparison of abstraction granularity
under identical pipeline conditions.

\subsection{Deriving Concrete Methods}
\label{sec:methods}

Each concrete memory method is an instantiation of a (scope, abstraction) pair.
Table~\ref{tab:method-derivation} shows how the four methods evaluated
in this paper are derived from the framework
(Appendix~\ref{sec:appendix-method-motivation} discusses the design
rationale for each).

\begin{table}[t]
\centering
\footnotesize
\caption{Memory methods as (scope, abstraction) instantiations.
Methods marked (ours) are novel; Reflection corresponds to
Reflexion~\citep{shinn2023reflexion} / LATS~\citep{zhou2024language}.}
\label{tab:method-derivation}
\begin{tabular}{@{}lll@{}}
\toprule
\textbf{Method} & \textbf{Scope} & \textbf{Abstraction} \\
\midrule
No Memory (baseline) & --- & --- \\
Raw Sibling (ours)  & within-expansion & raw \\
Reflection          & cross-trajectory    & reflection \\
LiTS-Fact    & cross-trajectory    & fact extraction \\
\bottomrule
\end{tabular}
\end{table}

Cross-sibling and cross-trajectory scopes are orthogonal:
Raw Sibling provides immediate diversity within a single expansion,
while LiTS-Fact provides accumulated knowledge across iterations.
Their composition (Eq.~\eqref{eq:combined}) tests whether
the two information sources are complementary:
\begin{equation}
\label{eq:combined}
a_i \sim \pi_\theta\!\bigl(a \mid s,\;
  \underbrace{\mathcal{C}_{\text{fact}}}_{\text{cross-traj}},\;
  \underbrace{\{(a_j, o_j)\}_{j<i}}_{\text{cross-sib}}
\bigr).
\end{equation}

\paragraph{Evaluated cells.}
The full $2 \times 3$ matrix of (scope, abstraction) pairs contains
six cells; three are degenerate or undefined
(Appendix~\ref{app:omitted-cells}).
The four methods in Table~\ref{tab:method-derivation} are the
well-defined, non-degenerate instantiations.
Not every method applies to every inference method: within-expansion
scope requires multi-candidate expansion (beam search, MCTS),
and cross-trajectory scope requires multiple iterations.
These constraints determine which cells of the experiment matrix
(Table~\ref{tab:experiment-matrix}) are admissible;
\S\ref{sec:task-types} gives the full admissibility rules.

\subsection{Task Types}
\label{sec:task-types}

Our study targets tool-use agents (\S\ref{sec:intro}): multi-step
interactive reasoning where the agent repeatedly issues structured
calls (SQL queries, SPARQL queries, shell commands) to an external
environment. This includes both single-tool multi-step settings
(an SQL agent issuing repeated queries to one database) and multi-tool
settings (a CLI agent invoking different shell commands).

\paragraph{Serializable vs.\ non-serializable environments.}
A serializable environment allows saving and restoring state
(e.g., a read-only database).
Tree search and beam search require serializability
because they explore multiple branches from the same state.
A non-serializable environment (e.g., a Docker container
whose filesystem is mutated by each command)
cannot be forked, restricting the agent to sequential trajectories.
Memory-augmented best-of-$N$ is therefore the only
multi-trajectory strategy available in non-serializable settings.

\paragraph{Admissibility of method-search combinations.}
The unified scope $\times$ abstraction $\times$ inference-method
framework determines which combinations are well-defined.
\textit{Cross-sibling scope} requires multi-candidate expansion (Beam,
MCTS) and is undefined for best-of-$N$;
\textit{cross-trajectory abstractions} (Reflection, Fact memory) require
multiple iterations and on 1-iteration Beam Search collapse to their
per-trajectory variants in the best-of-$N$ row;
\textit{within-trajectory injection} (per-step Fact) generalizes to
multi-iteration settings without modification.
These constraints carve the experimental matrix
(Table~\ref{tab:experiment-matrix}) into the cells we report
and explain the empty cells as inadmissible rather than untested.

%% file: sections/4.exp_setup.tex
\section{Experimental Setup}
\label{sec:setup}

\paragraph{Inference methods.}
\label{sec:inference-methods}
We consider three inference methods that differ in search structure.
\textbf{Best-of-$N$:} $N$ trajectories
are generated sequentially; memory transforms i.i.d.\ sampling into
sequential attempts conditioned on prior knowledge.
This is the only viable multi-trajectory strategy for
non-serializable environments~\citep{zainullina2025guided}
where state cannot be forked (e.g., Terminal-Bench).
\textbf{Beam search}~\citep{yao2023tree}\textbf{:} $B$ candidate
continuations are scored per step and the top-$B$ retained; memory
operates at per-step granularity across beams.
\textbf{MCTS}~\citep{hao-etal-2023-reasoning,zhou2024language}\textbf{:}
UCT selection, expansion, simulation, and backpropagation; memory
operates at both per-step (expansion) and per-trajectory
(backpropagation) granularity.
We follow Reflexion's policy-only injection convention across all
Reflection cells (Appendix~\ref{sec:appendix-injection}); the LATS
variant of injecting reflections into the reward prompt is available
under MCTS but is not used in our matrix to keep the per-method
comparison controlled.
Backpropagation configuration and full search settings are
documented in Appendix~\ref{sec:appendix-backprop}
and~\ref{sec:appendix-search-settings}.

\paragraph{Memory methods.}
\label{sec:memory-methods}
We evaluate the four methods in Table~\ref{tab:method-derivation}:
No Memory, Raw Sibling, Reflection, and LiTS-Fact; their
scope $\times$ abstraction construction is given in
\S\ref{sec:methods} and per-method details (triggers, applicability)
in Appendix~\ref{sec:appendix-methods-detail}.
LiTS-Fact adapts the atomic-fact extraction and deduplication
mechanism of mem0~\citep{chhikara2025mem0} (originally designed for
cross-session dialogue) to the cross-trajectory scope of
multi-attempt search, where the goal is caching reusable
environmental knowledge so that subsequent attempts skip redundant
discovery steps.

\paragraph{Benchmarks.}
\label{sec:benchmarks}
All benchmarks are tool-use tasks with structured tool calls.
\textbf{WikiSQL / WikiTQ} (SQL Agent) \citep{liu2024agentbench}: the agent queries a
relational database; serializable (read-only state).
\textbf{KGQA} (KG Agent) \citep{liu2024agentbench}: the agent traverses Freebase via SPARQL
for multi-hop questions; serializable.
\textbf{Terminal-Bench 2.0} (CLI Agent) \citep{merrill2026terminalbench}: the agent executes shell
commands to complete software engineering tasks;
\emph{non-serializable} (state is mutated and cannot be forked),
making beam search and MCTS infeasible.

\paragraph{Implementation.}
All experiments use LiTS~\citep{li2026lits}, a modular tree-search
framework whose policy, transition, and reward-model components are
shared across all configurations.
Memory is injected via a single abstraction that hooks into the
search loop without modifying other components
(formal pipeline in Appendix~\ref{sec:appendix-pipeline});
Raw Sibling additionally requires interleaved expansion ordering
(Eq.~\ref{eq:sibling-expand}).

\paragraph{Models and reporting.}
We use Claude Sonnet~4.6 as the policy on KGQA (entity resolution
and relation selection require strong language ability) and Claude
Haiku~3.5 on WikiSQL, WikiTQ, and Terminal-Bench (where Sonnet's
greedy baseline is near-ceiling, e.g., 76.5\% on WikiSQL, leaving
no headroom for memory experiments).
The per-step reward model and all augmentor LLMs are Sonnet~4.6
across every configuration, so memory effects within each row of
Table~\ref{tab:experiment-matrix} are isolated from supervisor-side
variation.
KGQA cells are reported on a common 69-example subset
(60 ReAct-error + 9 stratified-correct, seed 42;
Appendix~\ref{app:kgqa-subset});
all other benchmarks are reported on their full evaluation sets
(WikiSQL: 51, WikiTQ: 49, Terminal-Bench: 89).

\begin{table*}[t]
\centering
\footnotesize
\caption{Experiment matrix: memory method $\times$ inference method
$\times$ task type.
\textbf{Bold}: best accuracy within each inference-method group.
Significance markers: $**$ $p{<}0.01$, $*$ $p{<}0.05$ (McNemar's
exact test vs.\ the No-Memory baseline in the same row group);
raw p-values in Appendix~\ref{app:mcnemar-pvalues}.
Terminal-Bench is non-serializable (---).
Raw Sibling requires multi-candidate expansion (Beam/MCTS).
\textsuperscript{\dag}Not evaluated: LiTS-Fact is accuracy-neutral
across all other cells (F3); the additional KGQA MCTS run would
confirm the same null at high cost.}
\label{tab:experiment-matrix}
\begin{tabular}{@{}l l cccc@{}}
\toprule
\textbf{Inference Method} & \textbf{Memory Method}
  & \multicolumn{2}{c}{\textbf{SQL Agent}} & \textbf{KG Agent} & \textbf{CLI Agent} \\
  & & WikiSQL & WikiTQ & KGQA & Terminal-Bench \\
\midrule
\multicolumn{2}{@{}l}{\textbf{Single Trajectory} (ReAct, $T{=}0$)}
  & 39.2 & 28.6 & 13.0  & 7.9 \\
\midrule
\multirow{3}{*}{\textbf{Best-of-$N$} (pass@5, $T{=}0.9$)}
  & No Memory                & 49.0 & 40.8  & 31.9  & 23.6 \\
  & Reflection       & \textbf{58.8} & \textbf{49.0}  & \textbf{33.3}  & \textbf{25.8} \\
  & LiTS-Fact     & 47.1 & 34.7  & 31.9  & \textbf{25.8} \\
\midrule
\multirow{2}{*}{\textbf{Beam Search}}
  & No Memory                                     & 27.5 & 26.5 & 7.2 & --- \\
  & Raw Sibling (ours)                            & \textbf{39.2} & 30.6 & \textbf{27.5}$^{**}$ & --- \\
\midrule
\multirow{4}{*}{\textbf{MCTS}}
  & No Memory                & 49.0 & 38.8 & 23.2         & --- \\
  & Raw Sibling (ours)                            & 47.1 & 36.7 & \textbf{34.8}$^{*}$          & --- \\
  & Reflection                       & \textbf{60.8}$^{*}$ & \textbf{44.9} & \textbf{34.8}$^{*}$          & --- \\
  & LiTS-Fact              & 47.1 & 38.8 & \textsuperscript{\dag}             & --- \\
\bottomrule
\end{tabular}
\end{table*}

%% file: sections/5.results-and-analysis.tex
\section{Results and Analysis}
\label{sec:results}

We organise the results around three findings that emerge from the
experiment matrix in Table~\ref{tab:experiment-matrix}, supported by
McNemar's exact paired-binary test on each method-pair comparison
(Appendix~\ref{app:mcnemar-pvalues} lists raw p-values).%
\footnote{Throughout, we report McNemar's discordant pair counts as
($b$ rescues / $c$ regressions): a rescue is a baseline-incorrect
example that becomes correct under the treatment, a regression is
the reverse.}
\textbf{(F1)} Memory's accuracy effect is search-method-dependent:
neither Reflection nor Raw Sibling helps on every cell, and the cells
where each helps follow a coherent pattern (\S\ref{sec:f1-accuracy}).
\textbf{(F2)} On the harder benchmark (KGQA) under MCTS specifically,
the two otherwise-different memory abstractions become indistinguishable
in both their effect size and the specific examples they rescue;
this convergence does not hold on the SQL benchmarks, where Reflection
substantially outperforms Raw Sibling under MCTS, so we report it as
a single-cell finding rather than a generalisation
(\S\ref{sec:f2-convergence}).
\textbf{(F3)} Fact extraction is accuracy-neutral but
shortens trajectories by 19--26\% on tasks with reusable structure;
this is the only main result for which we observe a consistent
\emph{efficiency} effect rather than an accuracy effect
(\S\ref{sec:f3-efficiency}).

\subsection{F1: Memory's accuracy effect is search-method-dependent}
\label{sec:f1-accuracy}

The four method-pair comparisons that reach $p < 0.05$ under
McNemar's exact test (Table~\ref{tab:experiment-matrix}) cluster in
two patterns rather than spreading uniformly across cells:

\paragraph{Pattern 1 --- Reflection improves accuracy directionally
under both inference methods, but reaches significance only under
MCTS.}
Across all seven Reflection-vs-No-Memory cells where the comparison
is defined (Table~\ref{tab:experiment-matrix}), Reflection's effect
is directionally positive.
However, only two cells reach McNemar significance, and both lie in
the MCTS column: MCTS$+$Reflection on WikiSQL (6 rescues / 0
regressions, $p{=}0.031^{*}$) and on KGQA (10 rescues / 2
regressions, $p{=}0.039^{*}$).
The corresponding pass@5 cells on the same examples do not reach
significance: WikiSQL pass@5$+$Reflection has 5 rescues and 2
regressions ($p{=}0.453$), and KGQA pass@5$+$Reflection has 5
rescues and 4 regressions ($p{=}1.0$).
WikiTQ is consistent in direction in both columns but reaches
significance in neither (pass@5: 4 rescues / 1 regression,
$p{=}0.375$; MCTS: 5 rescues / 2 regressions, $p{=}0.453$); at
$b{+}c$ values of 5 and 7 the WikiTQ cells are plausibly
underpowered, so we read them as neutral rather than confirming or
rejecting the asymmetry.
On Terminal-Bench (Best-of-$N$ only) Reflection adds 8 rescues and
6 regressions ($p{=}0.791$); without an MCTS run the
within-benchmark contrast does not exist.
No cell exhibits the reverse pattern (a significant
pass@$N+$Reflection gain alongside a null or negative
MCTS$+$Reflection effect), so no cell rejects Pattern 1.

\paragraph{Mechanism: the per-step reward model filters bad
reflections.}
The most plausible explanation is that MCTS's per-step reward model
(PRM) evaluates each reflection-induced expansion and prunes paths
the reflection incorrectly recommends before they reach a terminal,
whereas under pass@$N$ the same bad reflection runs to completion
and shows up as a regression.
This filtering account predicts that, within a benchmark,
MCTS$+$Reflection should produce fewer regressions than
pass@5$+$Reflection.
The prediction holds on WikiSQL (0 vs.\ 2 regressions) and KGQA
(2 vs.\ 4) and is the proximate cause of MCTS$+$Reflection reaching
significance on these two cells; it does not hold on WikiTQ
(2 vs.\ 1), which we read as evidence that the PRM's filtering is
not uniformly effective across tasks rather than evidence that
filtering does not occur.
We further trace the per-iteration progression on the two WikiSQL
examples that are rescued by MCTS$+$Reflection but not by
Indep$+$Reflection (Appendix~\ref{app:pattern1-mechanism}).
Both show the same pattern: iteration~0 (when no reflection content
has yet been generated) fails on the same path that MCTS$+$No
Memory fails on, and iteration~1 onwards, after reflection content
becomes available, the PRM selects a sibling expansion the
reflection guides the policy towards. On KGQA the per-iteration
trace is less informative because the PRM evaluates each Freebase
relation choice in isolation, making iteration~0 q-values vary
substantially across runs; we discuss this case in
Appendix~\ref{app:case-study-cross-cell}.

\paragraph{Pattern 2 --- Raw Sibling helps where the policy is
diversity-starved.}
The strongest effect in our study is KGQA Beam$+$Raw Sibling: 27.5\%
vs.\ 7.2\% for Beam without memory ($\Delta{=}{+}20.3$pp,
17 rescues / 3 regressions, $p{=}0.003^{**}$).
At $n_\text{iters}{=}1$, Beam exposes a policy diversity collapse
(Appendix~\ref{app:beam-giveup}): at $T{=}0.7$ sampling, sibling
candidates frequently converge to the same action or apology,
leaving the reward model no diverse candidate to rank.
Raw Sibling's interleaved expansion (Eq.~\ref{eq:sibling-expand})
biases each next sibling away from prior siblings' actions, which
on KGQA's wide action space unlocks 17 rescues.
The same mechanism produces a $+$11.7pp gain on WikiSQL Beam
(8 rescues / 2 regressions, $p{=}0.109$) and $+$4.1pp on WikiTQ
Beam (3 / 1, $p{=}0.625$); under MCTS the gain disappears or
reverses ($-$1.9pp WikiSQL, $-$2.0pp WikiTQ; both NS), because
UCT-driven re-expansion already supplies sufficient action
variation.
Cross-sibling memory is therefore a remedy for low-diversity
configurations rather than a universal addition.

\subsection{F2: Memory abstractions converge on KGQA MCTS}
\label{sec:f2-convergence}

Pattern 1 from \S\ref{sec:f1-accuracy} extends to a comparison the
paper has not made before: under MCTS on KGQA, Reflection and Raw
Sibling produce \emph{the same accuracy and almost the same example
set}, despite operating on different memory scopes
(cross-trajectory vs.\ cross-sibling) with different abstraction
levels (per-trajectory diagnosis vs.\ raw action--observation
injection).
Both methods reach 34.8\% (24/69) on the 69-example subset, both
rescue exactly 10 ReAct-error examples, and both regress on exactly
2 examples; 9 of the 10 rescued examples are shared; each method uniquely
rescues exactly one example the other does not.
A direct paired test between the two methods has discordant counts
$b{=}c{=}1$, $p{=}1.0$ --- the two are statistically indistinguishable
in their accuracy distribution at $N{=}69$.

\paragraph{The convergence does not hold on the SQL benchmarks.}
Under MCTS on WikiSQL, Reflection reaches 60.8\% while Raw Sibling
reaches 47.1\% ($\Delta{=}13.7$pp); on WikiTQ the gap is 8.2pp.
F2 is therefore a single-cell finding.
A plausible explanation is that on KGQA's harder subset, any
non-i.i.d.\ context suffices for recovery, so the abstraction
choice stops mattering; on SQL where the policy is closer to
solving unaided, Reflection's targeted diagnosis makes a difference.

\paragraph{Why the convergence is interesting even as a single-cell
finding.}
Prior work evaluates a single abstraction per study; any paper that
ran only one method on KGQA-MCTS would have reported $+11.6$pp and
attributed the gain to its chosen abstraction.
The convergence shows that single-abstraction evaluations may
attribute accuracy gains to the wrong cause.

\subsection{F3: Fact extraction trades accuracy for efficiency}
\label{sec:f3-efficiency}

Across all six evaluated cells, LiTS-Fact's accuracy effect is
indistinguishable from zero under McNemar's exact test (all
$p \ge 0.250$; see Appendix~\ref{app:mcnemar-pvalues} for
per-cell counts).
We therefore do \emph{not} make an accuracy claim for LiTS-Fact.
A KGQA-specific failure-mode analysis (over-anchoring on prior facts,
procedural-instruction injection, and over-generalised negative facts)
is reported in Appendix~\ref{app:kgqa-fact-regression}.

\paragraph{Where Fact memory does have a measurable effect: trajectory length.}
LiTS-Fact reduces mean trajectory length from 6.1 to 4.9 steps on
WikiSQL ($-$20\%) and from 5.8 to 4.7 on WikiTQ ($-$19\%) by
allowing later attempts to skip the \texttt{sql\_db\_list\_tables}
discovery call: the skip rate in attempts 1--4 rises from 4--5\%
(No Memory) to 70--77\% (LiTS-Fact).
On KGQA, mean trajectory length drops from 8.9 to 6.6 steps
($-$26\%) for analogous reasons (cached relation discoveries).
Figure~\ref{fig:step-progression} and Table~\ref{tab:efficiency}
(Appendix~\ref{app:efficiency}) show the per-attempt progression
and full efficiency breakdown.
On KGQA, mean trajectory length drops from 8.9 to 6.6 steps
($-$26\%) for analogous reasons (cached relation discoveries).
On Terminal-Bench, where each task is run in a fresh ad-hoc Docker
container with no reusable structure across attempts, fact extraction
has no analogous mechanism for shortening trajectories: the extracted
facts cannot be reused to skip discovery steps because each task's
environment is unique.
Across these cells, Fact's effectiveness scales with the
\emph{environment regularity} (SQL $\approx$ KGQA $\gg$
Terminal-Bench), matching the abstraction's intended target:
extracted facts capture what the environment contains
(schemas, relations) and become useful only when those facts
generalise across attempts.

\paragraph{Reflection and Fact memory are not additive.}
When both augmentors are run together (Best-of-$N$\ row, last
sub-cell of WikiSQL/WikiTQ), pass@5 falls to 52.9\% / 46.9\% --- below
Reflection alone (58.8\% / 49.0\%) --- and the
\texttt{list\_tables}-skip rate falls to 20\% / 23\% --- below LiTS-Fact
alone (77\% / 70\%).
The composition therefore loses on both axes simultaneously.
A case study (Appendix~\ref{app:conflict}) shows the mechanism: the
reflection prompt issues an explicit step-by-step plan that overrides
the implicit shortcut Fact memory would offer, so the policy follows
the plan and ignores the cached facts.

\input{sections/fig-step-progression}

%% file: sections/fig-step-progression.tex
\begin{figure}[t]
\centering
\begin{tikzpicture}
\begin{axis}[
    width=0.85\columnwidth,
    height=4.5cm,
    xlabel={Attempt},
    ylabel={Mean steps},
    xtick={0,1,2,3,4},
    xticklabels={$a_0$,$a_1$,$a_2$,$a_3$,$a_4$},
    ymin=4, ymax=10,
    legend style={at={(0.5,-0.25)}, anchor=north, font=\small, legend columns=2},
    grid=major,
    grid style={dashed, gray!30},
]
\addplot[blue, thick, mark=square*, mark size=2pt]
    coordinates {(0,8.9) (1,9.0) (2,8.9) (3,8.8) (4,9.0)};
\addplot[red, thick, mark=triangle*, mark size=2.5pt]
    coordinates {(0,9.1) (1,6.5) (2,6.1) (3,5.7) (4,5.6)};
\addplot[blue, thick, dashed, mark=square, mark size=2pt]
    coordinates {(0,6.3) (1,6.0) (2,6.1) (3,6.0) (4,6.0)};
\addplot[red, thick, dashed, mark=triangle, mark size=2.5pt]
    coordinates {(0,5.8) (1,5.0) (2,4.8) (3,4.6) (4,4.4)};
\legend{KG: No Memory, KG: LiTS-Fact, SQL: No Memory, SQL: LiTS-Fact}
\end{axis}
\end{tikzpicture}
\caption{Mean trajectory length by attempt.
Baselines (blue) remain flat across attempts ---
each trajectory starts from scratch.
LiTS-Fact (red) decreases as accumulated facts
let the agent skip discovery steps.
The effect is stronger on KGQA (solid, $-$3.4 steps by $a_4$)
than SQL (dashed, $-$1.4 steps),
reflecting longer baseline trajectories on KGQA.}
\label{fig:step-progression}
\end{figure}
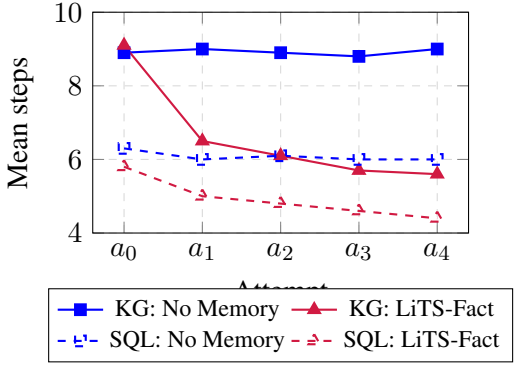

%% file: sections/6.discussion.tex
\section{Discussion}
\label{sec:discussion}

\paragraph{Why Reflection helps only under MCTS: comparison with prior work.}
LATS~\cite{zhou2024language} and Re-TRAC~\cite{zhu2026re} both
evaluate trajectory-level reflection with access to a binary
correctness signal (exact match, unit-test pass) during search.
Under such signals, it is unclear whether the reported gains
come from the reflection content or from the correctness signal
preventing bad trajectories from propagating.
Our verifier-free setup removes this confound: the only per-step
signal available during search is the LLM-based reward model (PRM).
The mechanism we propose (\S\ref{sec:f1-accuracy}) is that
Reflection and PRM act in concert --- Reflection provides
alternative paths informed by prior failures, and the PRM selects
among them --- which is why Reflection helps under MCTS (where PRM
operates) but not under best-of-$N$ (where no per-step selection
exists).

\paragraph{Connection to LLM sensitivity to irrelevant context.}
The Reflection$+$Fact composition failure
(\S\ref{sec:f3-efficiency}) is consistent with reports of LLMs
being sensitive to irrelevant context~\citep{shi2023large}: once an
explicit plan is present, the correct-but-redundant fact memory
becomes irrelevant and the policy ignores it, losing on both axes.

%% file: sections/7.conclusion.tex
\section{Conclusion}
\label{sec:conclusion}

We proposed a scope $\times$ abstraction taxonomy for cross-trajectory
memory in tool-use LLM agents and evaluated four methods under three
inference methods on four benchmarks in a controlled matrix.
The study identifies the inference method as a confound that must be
controlled for in any claim about a memory method's accuracy effect: the same memory
mechanism produces statistically distinct results under different
inference methods on the same examples.
Specifically, reflection reaches significance only under MCTS (not
best-of-$N$), cross-sibling injection helps only
diversity-starved Beam Search (not MCTS), and atomic fact extraction
is accuracy-neutral but shortens trajectories by 19--26\% on tasks
with reusable structure.
These findings suggest that published claims about cross-trajectory
memory effectiveness are not directly comparable across studies,
because each evaluates under a different inference method --- a
confound our controlled matrix makes explicit.

\subsection{Reproducibility}
We build on the open-source LiTS
framework~\citep{li2026lits}, extending it with a memory module
that implements the four methods in this paper.
We will release our fork (memory module + experiment configurations
+ evaluation scripts) upon acceptance.
Per-configuration evaluation results (per-example correctness
verdicts) and the analysis scripts used to produce all tables are
included in the supplementary material;
Appendix~\ref{app:compute-cost} reports API costs.

%% file: sections/app-rl-connection.tex
\section{Connection to Reinforcement Learning (Extended)}
\label{sec:appendix-rl-connection}

Our framework is an inference-time analogue of experience replay:
instead of storing raw $(s, a, r, s')$ transitions for gradient-based
policy updates, we store processed experience (reflections, extracted facts)
and inject them into the policy prompt.
The ``update'' is purely in-context --- no model parameters change.
This places our work in the broader family of in-context
learning approaches~\cite{brown2020gpt3}
that condition frozen LLMs on task demonstrations or interaction histories,
but differs in that we study \emph{what} to store (abstraction level)
and \emph{from where} to retrieve (memory scope)
rather than how to format demonstrations.
Hindsight Experience Replay~\cite{andrychowicz2017hindsight}
is a particularly close analogue:
it relabels failed trajectories with achieved goals
to extract learning signal from failures,
paralleling how our reflection augmentor
extracts strategy-level lessons from failed attempts.

%% file: sections/app-expansion.tex
\section{Batch vs.\ Sibling-Aware Expansion}
\label{sec:appendix-expansion}

When expanding a node with $N$ candidate actions,
the standard approach (\emph{batch expansion}) samples all candidates
independently from the same augmented policy:
\begin{equation}
\label{eq:batch-expand}
a_1, a_2, \ldots, a_N \overset{\text{i.i.d.}}{\sim} \pi_\theta(a \mid s, \mathcal{C}).
\end{equation}
This corresponds to the ``No Memory'' baseline in Table~\ref{tab:experiment-matrix}
(when $\mathcal{C} = \emptyset$) or to any cross-trajectory memory method
without cross-sibling scope (LATS Reflection, LiTS-Fact alone).

\emph{Sibling-aware expansion} (Eq.~\eqref{eq:sibling-expand} in \S\ref{sec:scope})
replaces i.i.d.\ sampling with sequential conditioning,
creating a dependent chain where each candidate sees
the actions and observations of previously sampled siblings.
This breaks the i.i.d.\ assumption but increases action diversity
at the cost of sequential latency.

%% file: sections/app-method-motivation.tex
\section{Method Motivation Details}
\label{sec:appendix-method-motivation}

This appendix expands the brief motivation in \S\ref{sec:methods}.

\paragraph{Why fact extraction?}
In tool-use tasks, agents repeatedly discover the same environmental
knowledge across trajectories
(e.g., querying database schemas, listing available API endpoints).
Fact extraction captures this knowledge as reusable atomic statements,
potentially saving redundant discovery steps in subsequent trajectories.
This is distinct from reflection, which captures strategy-level advice
but does not encode concrete environmental state.

\paragraph{Why cross-trajectory fact memory with best-of-$N$?}
best-of-$N$ are the only viable multi-trajectory strategy
for non-serializable environments~\citep{zainullina2025guided},
but without memory, each trajectory starts from scratch ---
repeating the same exploration and making the same mistakes.
Cross-trajectory fact memory transforms i.i.d.\ sampling
into sequential refinement:
each attempt inherits environmental knowledge
(discovered schemas, failed query patterns)
from prior attempts, without requiring state serialization.

\paragraph{Why cross-sibling raw context with tree search?}
In beam search and MCTS, multiple candidates are expanded
from the same node.
Without cross-sibling context, candidates are sampled i.i.d.\
from the same prompt, often producing identical actions
(action diversity collapse~\citep{li2026lits}).
Raw sibling context breaks this degeneracy at zero LLM cost:
each candidate sees what its predecessors already tried,
naturally diversifying the expansion without
requiring explicit diversity penalties or temperature tuning.

%% file: sections/app-omitted-cells.tex
\section{Omitted Cells in the Experiment Matrix}
\label{app:omitted-cells}

Cells are omitted from Table~\ref{tab:experiment-matrix} for two
distinct reasons. First, some (scope, abstraction) pairs in the
taxonomy are degenerate or undefined, so no concrete memory method
exists for them (\S\ref{app:degenerate-cells}). Second, some
well-defined memory methods are structurally incompatible with
certain inference methods (\S\ref{app:inadmissible-cells}).

\subsection{Degenerate scope \texorpdfstring{$\times$}{x} abstraction cells}
\label{app:degenerate-cells}
The $2 \times 3$ taxonomy (scope $\times$ abstraction) contains six
cells; three are not evaluated:
\begin{itemize}
\item \emph{Within-expansion $\times$ reflection}: reflection
  requires a complete trajectory to diagnose. Within a single
  expansion, each sibling has produced only one action--observation
  pair --- insufficient input for a trajectory-level summary.
\item \emph{Within-expansion $\times$ fact extraction}: within a
  single expansion, each sibling's observation is a single tool
  return (e.g., one SQL result set, one SPARQL entity list).
  Extracting ``atomic facts'' from a single observation yields the
  observation itself --- there is nothing to compress or
  deduplicate.
  Fact extraction's value lies in compressing \emph{long}
  multi-step trajectories (8--15 observations) into a small set of
  reusable statements; at the single-observation granularity of
  within-expansion scope, this compression is vacuous and the cell
  collapses to within-expansion $\times$ raw.
\item \emph{Cross-trajectory $\times$ raw}: injecting the full raw
  history of prior trajectories into the prompt is infeasible for
  long tool-use trajectories (8--15 steps $\times$ $\sim$500 tokens
  each $\times$ up to 5 prior trajectories exceeds typical context
  budgets). Reflection and fact extraction exist precisely to
  compress this history into a manageable representation.
\end{itemize}

\subsection{Inadmissible memory \texorpdfstring{$\times$}{x} inference-method cells}
\label{app:inadmissible-cells}

\paragraph{Terminal-Bench $\times$ Beam Search / MCTS (---).}
Terminal-Bench uses Docker containers whose state is mutated
by each shell command and cannot be forked.
Beam search and MCTS require state serialization
to explore multiple branches from the same node,
making them structurally infeasible in this environment
(\S\ref{sec:task-types}).

\paragraph{Beam Search $\times$ Cross-Trajectory Memory.}
Our beam search runs a single round of expansion and selection
(no further iteration over the resulting beam).
Cross-trajectory memory (reflection, fact extraction) generates context
only after a complete trajectory finishes, so any benefit would accrue
to a subsequent iteration; with no subsequent iteration, these cells are
structurally equivalent to No Memory.

\paragraph{KGQA $\times$ Fact + Reflection.}
On the 69-subset, KGQA Indep$+$Reflection is essentially flat
(31.9\% $\to$ 33.3\%, 5 rescues / 4 regressions, $p{=}1.0$;
\S\ref{sec:f1-accuracy}).
Adding reflection to fact extraction therefore cannot improve over
fact extraction alone --- it can only add cost or introduce the
abstraction-level conflict observed on WikiSQL/WikiTQ
(Appendix~\ref{app:conflict}).
We omit this combination.

\paragraph{Raw Sibling $\times$ best-of-$N$ ($\varnothing$).}
Raw Sibling requires multi-candidate expansion at the same node.
best-of-$N$ generate one trajectory at a time
with no shared expansion step, making cross-sibling context inapplicable.

%% file: sections/app-backprop.tex
\section{Backpropagation Configuration}
\label{sec:appendix-backprop}

The backpropagation mode in our MCTS rows is determined by two
factors: reward source and cross-rollout aggregation.

\paragraph{Reward source.}
All our benchmarks are tool-use tasks where no objective terminal
signal is available (unlike game-playing MCTS where a win/loss
outcome is propagated). Each node maintains its own per-step LLM
reward aggregated from its position to the leaf.

\paragraph{Cross-rollout aggregation.}
For non-memory baselines (standard MCTS), rollouts are i.i.d., so
we use cumulative backpropagation: each rollout appends a value and
$Q$ is the running mean.
For memory-augmented configurations, successive iterations are
non-independent (later rollouts benefit from accumulated memory
context), so we adopt the decay-based update from
\citet{zhang-etal-2025-llama}:
$Q(\text{parent}) \leftarrow (1{-}\gamma)\,Q(\text{parent})
+ \gamma\,Q(\text{child})$,
which assigns higher weight to more recent, better-informed
trajectories.

%% file: sections/app-search-settings.tex
\section{Search Settings: Adapting Reflexion and LATS to Verifier-Free Tool Use}
\label{sec:appendix-search-settings}

This appendix documents how the \textbf{Reflection} cells of
Table~\ref{tab:experiment-matrix} are configured under our verifier-free
deployment-faithful setup. Two prior methods are direct precursors:
Reflexion \citep{shinn2023reflexion} for the best-of-$N$ cell, and
LATS \citep{zhou2024language} for the beam-search and MCTS cells. We
share their core mechanism --- a per-trajectory LLM reflection that is
stored in a persistent buffer and injected into subsequent
trajectories --- but adapt three settings to a setup without an inline
binary task verifier. The same three adaptations apply to both
precursors, so we describe them once.

\subsection{Setting Choices Compared with Reflexion and LATS}

\begin{table*}[ht]
\centering
\footnotesize
\caption{Settings used for the Reflection cells, compared with
Reflexion (best-of-$N$) and LATS (tree search).}
\label{tab:lats-vs-ours}
\begin{tabular}{@{}lp{3.8cm}p{3.8cm}p{3.8cm}@{}}
\toprule
\textbf{Aspect} & \textbf{Reflexion} & \textbf{LATS} & \textbf{Ours} \\
\midrule
Inference & ReAct trials & MCTS / beam & both \\
Trigger signal & EM vs.\ gold (HotpotQA), heuristics (AlfWorld), unit tests (HumanEval) & EM / unit tests & continuous PRM threshold ($r{<}0.3$) \\
Stop-on-correct & yes (Evaluator passes) & yes (binary verifier) & no \\
Reflection injection & policy prompt & policy and reward prompts & policy prompt only \\
Reflection LLM & free choice ($M_{sr}$ separate) & not specified & free choice (Sonnet on Haiku) \\
Iteration / trial budget & loop until pass or max & search until solved & fixed budget \\
\bottomrule
\end{tabular}
\end{table*}

\paragraph{Inheritance.}
The mechanism in our Reflection cells is unchanged from these
precursors: a trajectory-level reflection is generated after a failed
trial, stored persistently, and injected into subsequent trials.
Reflexion explicitly allows its self-reflection model $M_{sr}$ to be a
distinct LLM from the actor $M_a$~\citep{shinn2023reflexion}, so our use of a
stronger augmentor (Sonnet) on top of a weaker policy (Haiku) is
within the published Reflexion framework, not a deviation.

\subsection{No Stop-on-Correct under a Verifier-Free Setup}
\label{sec:appendix-stop}

Both Reflexion and LATS terminate the search the first time a
candidate trajectory is judged correct by a binary signal: exact match
against gold answers (HotpotQA), unit-test pass (HumanEval), or
task-specific heuristics (AlfWorld). Our tool-use benchmarks
(SQL, KG, CLI) do not expose such a signal during search: SQL
execution-match against the gold answer, SPARQL answer-match, and
Terminal-Bench test suites are evaluation-time signals computed by
\textsc{lits-eval}, not search-time signals available to the agent.
Wiring these signals into the search loop would require the agent to
access dataset annotations (gold answers, gold SPARQL outputs, hidden
test cases) that are unavailable at deployment, undermining the
benchmark's deployment-faithfulness.

The remaining option --- stop-on-correct gated by the LLM-based PRM
that already guides expansion --- is incompatible with the goals of
this study for two reasons.

\textit{Self-validating signal.} If the same PRM both (a) ranks
candidate continuations during expansion and (b) decides when the
search has succeeded, the search stops the moment the PRM's value
estimate clears its own threshold, regardless of whether the answer is
actually correct. The benefit of the search --- averaging out PRM
noise across multiple candidate trajectories --- is forfeited
precisely when a noisy false-positive estimate triggers early
termination.

\textit{Mechanism observation.} The contribution of this paper is
mechanism-level: we study how cross-trajectory memory (reflection,
fact extraction) accumulates across multiple completed trajectories
within a single search. Stop-on-correct at the first attempt collapses
the search into a single-trajectory regime in which neither reflection
nor fact extraction has the opportunity to be triggered, retrieved, or
composed. The resulting accuracy comparison would not reflect the
intended experimental variable.

We therefore disable stop-on-correct and run a fixed iteration / trial
budget (\S\ref{sec:appendix-budget}) on every example, deferring
verification to \textsc{lits-eval} after the search.

\subsection{Continuous PRM Trigger}
\label{sec:appendix-trigger}

For the same reason --- no inline gold signal --- we cannot use
Reflexion's binary EM trigger or LATS's binary verifier trigger to
gate reflection generation. We trigger reflection on the same PRM
score used for search guidance: trajectories with $r < 0.3$ generate
reflections; higher-scored trajectories do not. The threshold is set
once and held fixed across all benchmarks. This converts the binary
trigger into a continuous one, consistent with the rest of our
verifier-free setup.

\subsection{Fixed Iteration Budget}
\label{sec:appendix-budget}

In place of stop-on-correct we use a fixed budget $N_{\text{budget}}$,
i.e., $n_\text{iters}{=}N_{\text{budget}}$ for MCTS and beam, or
$N_{\text{budget}}$ pass@$N$ trials for best-of-$N$. A
fixed budget is compute-bounded (per-example cost is predictable and
capped) and orthogonal to the PRM's value estimate, avoiding the
self-validation issue above. It preserves a multi-trajectory regime
in which cross-trajectory memory has the opportunity to act.

The budget value is chosen as a compromise between two pressures.
A small budget (e.g., $N_{\text{budget}}{=}1$) recovers a
single-trajectory regime that defeats the purpose of using a
multi-trajectory method at all. A large budget (e.g.,
$N_{\text{budget}}{=}30$) maximizes mechanism observability but
multiplies cost roughly $N_{\text{budget}}\times$, dominated by
per-iteration PRM evaluation. We use $N_{\text{budget}}{=}5$:
enough trajectories per example to expose memory dynamics while
keeping per-example cost within an order of magnitude of
best-of-$N$ pass@5.

\subsection{Reflection Injection Site}
\label{sec:appendix-injection}

The original LATS injects reflections from failed prior trajectories
into both the policy prompt (steering the next rollout) and the value
prompt (informing the reward model's scoring of the current
trajectory). Reflexion injects only into the policy prompt. We follow
Reflexion's convention --- injection into the policy prompt only ---
across all Reflection cells in our matrix.

The reason is a controlled comparison with LiTS-Fact: under our
framework, the two methods share an identical pipeline
(per-trajectory trigger, persistent storage, retrieval-and-injection
into the policy prompt) and differ only in \emph{what the LLM produces}
--- strategy-level advice versus observation-level atomic facts.
Adding reflection-into-reward to one method but not the other would
entangle two variables (abstraction granularity and number of injection
sites) and break the controlled study.

A variant injecting reflections into both the policy and the reward
prompt (as in the original LATS) is not evaluated in the main matrix.
This variant studies a complementary question --- whether providing
reflection text as additional context to the reward model yields
further gains beyond reflection-into-policy alone --- which is
orthogonal to the abstraction-granularity comparison that motivates
the main matrix.

\subsection{Why these Differences are not Confounds}

The three operational adaptations (continuous PRM trigger, no
stop-on-correct, policy-only injection) shift the cell uniformly
across all four memory methods evaluated under each inference strategy:
No Memory, Raw Sibling (where applicable), Reflection, LiTS-Fact,
and Fact~$+$~Reflection all use the same trigger, the same budget, and
the same injection site. Memory effects reported in
Table~\ref{tab:experiment-matrix} are measured relative to the No Memory
baseline under the same protocol, not relative to Reflexion's or LATS's
published numbers, and are not confounded by these protocol changes.
For the same reason, our \textbf{best-of-$N$ $\times$ Reflection} cell can be read directly as the closest correspondence to
Reflexion in our matrix, with the three adaptations above as the only
differences.

\subsection{Hyperparameter Summary}
\label{sec:appendix-hyperparams}

The full set of search hyperparameters used in the Reflection
cells of Table~\ref{tab:experiment-matrix}:
$n_\text{actions}{=}3$, $n_\text{iters}{=}5$, exploration constant
$w_\text{exp}{=}1.0$, simulate-strategy $=\max$, maximum step depth
$=15$, rollout depth $=15$. Memory-augmented runs use cumulative
backpropagation; the decay variant (\S\ref{sec:appendix-backprop}) is reserved
for runs where prior iterations' value estimates are explicitly
de-weighted in favor of memory-informed later iterations.

\subsection{Asymmetric Policy / Supervision Models}
\label{sec:appendix-models}

Throughout the WikiSQL and WikiTQ MCTS rows of
Table~\ref{tab:experiment-matrix} we use
Haiku~3.5 as the policy LLM and Sonnet~4.6 for all
``supervision-side'' roles: the reward model, the reflection
generator, and the fact extractor.
On KGQA, Sonnet~4.6 serves as both policy and supervisor
(entity resolution and relation selection require stronger language
ability; see \S\ref{sec:setup}). The asymmetry --- weaker model for
action generation, stronger model for evaluation and memory
distillation --- is deliberate and consistent with Reflexion's
modular framework, in which the actor $M_a$ and the self-reflection
model $M_{sr}$ are explicitly allowed to differ \citep{shinn2023reflexion}.

\paragraph{Why this asymmetry does not confound the memory deltas.}
A stronger supervisor evaluating a weaker policy raises the absolute
accuracy level reported in the MCTS row, since Sonnet's reward signal
is more permissive than Haiku's would be. Memory effects in
Table~\ref{tab:experiment-matrix} are reported as deltas from a No
Memory baseline that uses the \emph{same} Sonnet supervisor across
the row (\S\ref{sec:appendix-models} below). The level inflation is
identical across the No Memory and memory-augmented cells, so it
cancels in any pairwise comparison; what remains is the contribution
of the memory mechanism itself.

%% file: sections/app-methods-detail.tex
\section{Memory Methods: Per-Method Details}
\label{sec:appendix-methods-detail}

We give the per-method specification used in our experiments.

\paragraph{No Memory (baseline).}
Standard inference without any cross-sibling or cross-trajectory
information sharing.
In beam search and MCTS, candidates at the same node are sampled
i.i.d.\ from the same prompt
(batch expansion, Eq.~\eqref{eq:augmented-policy} with
$\mathcal{C} = \emptyset$).

\paragraph{Raw Sibling (ours).}
Cross-sibling scope with raw abstraction.
Interleaved expansion (Eq.~\eqref{eq:sibling-expand}):
each candidate sees the complete action--observation pairs
of previously sampled siblings, injected as a diversity prompt.
No LLM summarization; zero additional LLM calls.
Only applicable to beam search and MCTS.

\paragraph{LATS Reflection~\citep{zhou2024language}.}
Cross-trajectory scope with reflection abstraction.
After each trajectory completes, an LLM generates a strategy-level summary.
Trigger: per-trajectory.
In MCTS, reflections can additionally be injected into the reward prompt
(denoted ``+reward'' in Table~\ref{tab:experiment-matrix}).

\paragraph{LiTS-Fact.}
Cross-trajectory scope with atomic fact extraction,
adapted from mem0~\citep{chhikara2025mem0} for the multi-attempt setting.
An LLM extracts discrete factual statements from observations.
Facts are deduplicated via embedding-based cosine similarity
at write time; at inference, all facts from prior attempts
are injected into the policy prompt.
Trigger: per-step in beam search and MCTS (incremental extraction);
per-trajectory in best-of-$N$
(batch extraction from the full trajectory in a single LLM call).

\paragraph{Raw Sibling + LiTS-Fact (ours).}
Composition of cross-sibling and cross-trajectory scopes
(Eq.~\eqref{eq:combined}).
Tests whether immediate sibling diversity and accumulated
cross-trajectory knowledge are complementary.
Only applicable to beam search and MCTS.

%% file: sections/app-pipeline.tex
\section{Augmentor Pipeline (Formal)}
\label{sec:appendix-pipeline}

Each augmentor follows a four-stage pipeline:
\begin{enumerate}
\item \textbf{Invocation}: an event in the search loop triggers the augmentor
  (after a step completes, after a trajectory terminates).
\item \textbf{Analysis}: $f_k(\mathcal{H}_k)$ produces a context unit~$c$
  (a textual insight, structured issue, or factual memory).
\item \textbf{Persistence decision}: a predicate $g(c) \in \{0, 1\}$ determines
  whether $c$ is stored in a persistent memory $\mathcal{M}$:
\begin{equation}
\mathcal{M} \leftarrow \mathcal{M} \cup \{c \mid g(c) = 1\}.
\end{equation}
  Raw Sibling context is ephemeral ($g=0$; exists only during the current expansion).
  Reflection and fact extraction are persistent ($g=1$; accumulated across iterations).
\item \textbf{Retrieval and injection}: before action sampling,
  relevant context is retrieved from $\mathcal{M}$
  (and any ephemeral context from stage~2)
  and injected into the policy prompt.
\end{enumerate}
This pipeline unifies all four methods in
Table~\ref{tab:method-derivation} as instances of the same
augmentor interface with different choices of scope, abstraction,
invocation point, and persistence predicate.

%% file: sections/app-kgqa-subset.tex
\section{KGQA Evaluation Subset Selection}
\label{app:kgqa-subset}

For KGQA, tree search methods (Beam Search, MCTS) require a per-step
reward model call using Sonnet~4.6 (\$3/\$15 per 1M input/output tokens),
making full 150-example evaluation cost-prohibitive
(\textasciitilde\$2.90/example $\times$ 150 = \textasciitilde\$435
per configuration, measured from a completed pilot;
5 configurations = \textasciitilde\$2,175 total).
To enable direct cross-method comparison within a feasible budget,
we evaluate all methods---including best-of-$N$---on
a common 69-example stratified subset.

\paragraph{Subset construction.}
Starting from the 150-example KGQA dataset (GrailQA subset from AgentBench),
we construct the evaluation subset in two steps:
\begin{enumerate}
\item \textbf{Error examples (60):}
  All examples where the ReAct baseline (Sonnet~4.6, greedy $T{=}0$)
  produces an incorrect answer (F1 $< 1.0$).
  These are the examples where tree search and memory have the
  opportunity to rescue.
\item \textbf{Stratified-correct controls (9):}
  From the 90 ReAct-correct examples, we draw a stratified random
  sample of 10 (one per decile of the sorted index range),
  using fixed seed 42 for reproducibility.
  One sampled index (idx=1) overlaps with the initial pilot run
  and is excluded from the subset command, yielding 9 controls.
  These validate the assumption that search methods do not regress
  on examples the baseline already solves.
\end{enumerate}

\paragraph{Reproducibility.}
The subset indices are deterministically generated via stratified
random sampling (seed 42) and are provided in our code release.
All KGQA configurations in Table~\ref{tab:experiment-matrix}
--- including best-of-$N$ baselines ---
are evaluated on this identical 69-example set.
We report accuracy as the fraction correct within the subset.

%% file: sections/app-mcnemar-pvalues.tex
\section{McNemar's Exact Test: Raw P-Values}
\label{app:mcnemar-pvalues}

For every method-pair comparison the paper makes, we report McNemar's
two-sided exact binomial test on the discordant pairs.
For paired binary outcomes, $b$ rescued (B correct, A wrong) and $c$
regressed (A correct, B wrong) under the null
$P(\text{rescue} \mid \text{discordant}) = 0.5$ follow
$\text{Binomial}(b{+}c, 0.5)$; the exact two-sided $p$-value is
$2 \cdot \sum_{k{=}0}^{\min(b,c)} \binom{b{+}c}{k} 0.5^{b{+}c}$,
capped at 1.
The exact form is required because several discordant counts in our
study fall below the $b{+}c \gtrsim 25$ threshold for the standard
chi-square approximation.
The full table is reproducible via
\texttt{paper/lits\_memory/results/scripts/mcnemar\_test.py}.
Significance markers: $^{**}$ $p{<}0.01$, $^{*}$ $p{<}0.05$,
$^{\dagger}$ $p{<}0.10$.

\begin{table*}[ht]
\centering
\small
\caption{McNemar's exact test results for every method-pair comparison
the paper claims, grouped by benchmark.
$N$ is the number of examples on which both methods ran.
$A_\text{corr}$ and $B_\text{corr}$ are the per-method accuracy counts
on those $N$ examples; \emph{res} is the count of examples where B is
correct and A is wrong (rescues), \emph{reg} the reverse (regressions).}
\label{tab:mcnemar-pvalues}
\begin{tabular}{@{}llrrrrrl@{}}
\toprule
\textbf{Benchmark} & \textbf{Comparison (A vs.\ B)}
  & $N$ & $A_\text{corr}$ & $B_\text{corr}$ & res & reg & $p$ \\
\midrule
\multirow{7}{*}{\textbf{WikiSQL}}
& Best-of-$N$: Refl vs.\ No Mem               & 51 & 25 & 30 & 5 & 2 & 0.453 \\
& Best-of-$N$: Fact vs.\ No Mem               & 43 & 20 & 19 & 2 & 3 & 1.000 \\
& Beam: Raw Sib vs.\ No Mem                  & 51 & 14 & 20 & 8 & 2 & 0.109 \\
& MCTS: Refl vs.\ No Mem                     & 51 & 25 & 31 & 6 & 0 & $0.031^{*}$ \\
& MCTS: Fact vs.\ No Mem                     & 51 & 25 & 24 & 0 & 1 & 1.000 \\
& MCTS: Raw Sib vs.\ No Mem                  & 51 & 25 & 24 & 2 & 3 & 1.000 \\
& MCTS$+$Refl vs.\ pass@5$+$Refl             & 51 & 27 & 31 & 6 & 2 & 0.289 \\
\midrule
\multirow{7}{*}{\textbf{WikiTQ}}
& Best-of-$N$: Refl vs.\ No Mem               & 48 & 19 & 22 & 4 & 1 & 0.375 \\
& Best-of-$N$: Fact vs.\ No Mem               & 48 & 19 & 16 & 0 & 3 & 0.250 \\
& Beam: Raw Sib vs.\ No Mem                  & 49 & 13 & 15 & 3 & 1 & 0.625 \\
& MCTS: Refl vs.\ No Mem                     & 49 & 19 & 22 & 5 & 2 & 0.453 \\
& MCTS: Fact vs.\ No Mem                     & 49 & 19 & 19 & 2 & 2 & 1.000 \\
& MCTS: Raw Sib vs.\ No Mem                  & 49 & 19 & 18 & 2 & 3 & 1.000 \\
& MCTS$+$Refl vs.\ pass@5$+$Refl             & 49 & 23 & 22 & 1 & 2 & 1.000 \\
\midrule
\multirow{7}{*}{\textbf{KGQA (69-subset)}}
& Best-of-$N$: Refl vs.\ No Mem               & 69 & 22 & 23 & 5 & 4 & 1.000 \\
& Best-of-$N$: Fact vs.\ No Mem               & 69 & 22 & 22 & 4 & 4 & 1.000 \\
& Beam: Raw Sib vs.\ No Mem                  & 69 &  5 & 19 & 17 & 3 & $0.003^{**}$ \\
& MCTS: Refl vs.\ No Mem                     & 69 & 16 & 24 & 10 & 2 & $0.039^{*}$ \\
& MCTS: Raw Sib vs.\ No Mem                  & 69 & 16 & 24 & 10 & 2 & $0.039^{*}$ \\
& MCTS$+$Refl vs.\ pass@5$+$Refl             & 69 & 23 & 24 & 6 & 5 & 1.000 \\
& MCTS$+$Refl vs.\ MCTS$+$Raw Sib            & 69 & 24 & 24 & 1 & 1 & 1.000 \\
\midrule
\multirow{2}{*}{\textbf{Terminal-Bench}}
& Best-of-$N$: Refl vs.\ No Mem               & 89 & 21 & 23 & 8 & 6 & 0.791 \\
& Best-of-$N$: Fact vs.\ No Mem               & 89 & 21 & 23 & 5 & 3 & 0.727 \\
\bottomrule
\end{tabular}
\end{table*}

\paragraph{Multiple testing.}
We do not apply a family-wise correction (e.g., Bonferroni) because
several of our headline tests are pre-specified by the unified
framework (\S\ref{sec:framework}) rather than chosen post-hoc, and
because Bonferroni assumes independent tests, a poor fit for our
correlated comparisons (the same memory method evaluated on related
benchmarks).
Benjamini-Hochberg FDR control is more appropriate for our setting;
applying it at $q{=}0.05$ to the 23-comparison family above leaves
KGQA Beam Raw Sib (uncorrected $p{=}0.003$) at the boundary of the BH
threshold, with the three other $p{<}0.05$ findings just above.
We discuss this caveat in \S\ref{sec:discussion} and report all
findings as $p{<}0.05$ uncorrected; readers can re-derive
FDR-adjusted p-values from the table above.

%% file: sections/app-pattern1-mechanism.tex
\section{Per-iteration Trace of Pattern 1: Reflection-driven MCTS Rescues}
\label{app:pattern1-mechanism}

This appendix traces the per-iteration progression of MCTS+Reflection
versus MCTS+No-Memory on the two WikiSQL examples that are rescued by
MCTS+Reflection but not by Indep+Reflection (\S\ref{sec:f1-accuracy},
Pattern 1). Both examples exhibit the same pattern: iteration~0
(reflection-free, since reflection content is generated only after
the first simulated trajectory) fails on the same path that MCTS
without memory fails on; iteration~1 onwards, after reflection
content is available,
the search finds a high-reward path. This is consistent with the
filtering account in \S\ref{sec:f1-accuracy}: the per-step reward
model selects the reflection-induced expansion among the candidate
siblings, while pass@$N$ has no analogous selection step within an
attempt.

\subsection{Case study A: WikiSQL example 26}
\label{app:case-study-a-26}

\paragraph{Question.}
\emph{``Name the platform for year more than 2006 and developer of
3g studios.''} The intended table is named
\texttt{SWAT~Games}; common-sense table names like
\texttt{game\_results} and \texttt{Games} exist in the database but
do not contain a \texttt{Developer} column.

\paragraph{Pass@5+Reflection (Indep+Refl).}
All five attempts apologise without identifying \texttt{SWAT~Games},
even though attempts 2--4 see reflections from earlier attempts in
their context (the Reflection memory diagnoses
\emph{``find a table with platform, year, developer columns''} after
attempt~0). Attempts 2 and 3 list tables and inspect schemas of
\texttt{game\_results} / \texttt{game~score}; both lack
\texttt{Developer}. The trajectory ends in the apology terminal
without ever inspecting \texttt{SWAT~Games}'s schema.

\paragraph{MCTS+No-Memory.}
Across iterations 0--4 the best q-mean stays in the
0.68--0.71 range; the best path's last action remains
\texttt{sql\_db\_list\_tables} or \texttt{sql\_db\_schema(game\_results)}
in every iteration. The agent never advances to a SELECT against
\texttt{SWAT~Games}.

\paragraph{MCTS+Reflection.}
Iteration~0 receives no reflection content (the execution log
records \texttt{ReflectionAugmentor returned empty for
traj\_key=q/0/...}, since reflection units are generated only after
a trajectory completes). It therefore behaves the same way as
MCTS+No-Memory: best q-mean is~0.66, best path stops at
\texttt{sql\_db\_list\_tables}. The first reflection
unit is then generated, diagnosing \emph{``wrong tables; need a
table with platform, year, developer columns.''}
At iteration~1, the policy expands a sibling that issues
\texttt{sql\_db\_schema(table\_names="game\_results, SWAT Games")} ---
the first time \texttt{SWAT~Games} appears in any candidate. The
PRM scores the schema-call branch high; the subsequent SELECT
against \texttt{SWAT~Games} is simulated to terminal at q-mean
\textbf{0.934}. Iterations 2--4 refine the SQL syntax; the q-mean
stays in the 0.91--0.93 range. Table~\ref{tab:case26-iters}
summarises the per-iteration trace.

\begin{table}[t]
\footnotesize
\centering
\caption{Per-iteration best q-mean, WikiSQL example~26.
MCTS+Refl jumps at iter~1 (first reflection-available iteration);
MCTS+No-Mem plateaus.}
\label{tab:case26-iters}
\begin{tabular}{@{}l rrrrr@{}}
\toprule
\textbf{iter} & 0 & 1 & 2 & 3 & 4 \\
\midrule
MCTS$+$No Mem   & 0.68 & 0.71 & 0.69 & 0.69 & 0.69 \\
MCTS$+$Refl  & 0.66 & \textbf{0.93} & 0.92 & 0.92 & 0.93 \\
\bottomrule
\end{tabular}
\end{table}

\subsection{Case study B: WikiSQL example 29}
\label{app:case-study-b-29}

\paragraph{Question.}
\emph{``What is the earliest game with a score of 99-89?''}
The intended table is \texttt{Game~Results} (capitalised, with a
space); the look-alike \texttt{game\_results} table lacks the
required date format.

\paragraph{Trace.}
The same pattern repeats. MCTS+No-Memory's best q stays in
0.70--0.73 across iterations 0--4; the agent loops between
\texttt{game\_results} and \texttt{game~score} without finding the
correct table. MCTS+Reflection's iteration~0 also fails (q$=$0.70)
on \texttt{game\_results}; reflection diagnoses \emph{``score format
mismatch; check distinct score formats''}; iterations 1--3 refine
score-format guesses; iteration~4 reaches \texttt{Game~Results}
through a sibling that the PRM scores high, q-mean reaches
\textbf{0.89}. The mechanism is the same as case~A but with more
iterations needed: each subsequent reflection refines the
specific format guess, and the PRM filters the
non-promising format-guess paths along the way.
Table~\ref{tab:case29-iters} summarises.

\begin{table}[t]
\footnotesize
\centering
\caption{Per-iteration best q-mean, WikiSQL example~29.
MCTS+Refl's correct path emerges at iter~4 after multiple reflection
refinements; MCTS+No-Mem plateaus throughout.}
\label{tab:case29-iters}
\begin{tabular}{@{}l rrrrr@{}}
\toprule
\textbf{iter} & 0 & 1 & 2 & 3 & 4 \\
\midrule
MCTS$+$No Mem   & 0.73 & 0.70 & 0.70 & 0.73 & 0.70 \\
MCTS$+$Refl  & 0.70 & 0.70 & 0.73 & 0.70 & \textbf{0.89} \\
\bottomrule
\end{tabular}
\end{table}

\subsection{Cross-cell summary and KGQA caveat}
\label{app:case-study-cross-cell}

The two case studies above show the iter-0 / iter-1+ asymmetry
clearly because WikiSQL has a strong floor effect: when the policy
picks the wrong table, all five iterations get stuck at the same
plateau, so any non-zero progress in MCTS+Reflection is attributable
to the reflection content that became available after iteration~0.

KGQA admits four examples that are rescued by MCTS+Reflection but
not by Indep+Reflection (idx~3, 37, 46, 110). The per-iteration
trace on these examples is less informative for mechanism
attribution: KGQA has a denser per-step reward signal (the PRM
evaluates each Freebase relation choice in isolation), so
iteration~0 q-values already vary substantially across runs. On
all four KGQA cases, MCTS+Reflection's iteration~0 best q-mean is
within 0.02--0.07 of the final iteration~4 q-mean, while
MCTS+No-Memory's iteration~0 q-mean is correspondingly lower than
its iteration~4 by a similar margin. The aggregate effect (the
$+11.6$pp McNemar-significant gain on the 69-subset) is real, but
the per-example mechanism on KGQA is best described as accumulated
variance reduction over five MCTS iterations rather than a single
reflection-driven rescue moment per example.

We therefore report Pattern~1's mechanism as confirmed at the
per-iteration level on WikiSQL (where iter~0 reliably fails) and as
aggregate-level only on KGQA. WikiTQ admits no examples in this
category and is therefore not analysed at the case-study level.

%% file: sections/app-beam-giveup.tex
\section{Beam Search Give-Up Analysis}
\label{app:beam-giveup}

On WikiSQL (51 examples) and WikiTQ (49 examples), beam search
(beam size 3, Haiku~3.5 policy at $T{=}0.7$, Sonnet~4.6 PRM)
underperforms greedy ReAct ($T{=}0$).
Table~\ref{tab:beam-giveup} summarizes the give-up phenomenon.

\begin{table}[ht]
\centering
\footnotesize
\caption{Beam search give-up analysis.
\emph{Apol.\ terminals}: terminals with ``I cannot find...'' answers.
\emph{Sel.\ apol.}: examples where the best terminal is an apology.
\emph{All-apol.\ states}: expand states where all 3 siblings apologize.}
\label{tab:beam-giveup}
\resizebox{\columnwidth}{!}{%
\begin{tabular}{@{}l cc cc@{}}
\toprule
& \multicolumn{2}{c}{\textbf{WikiSQL}} & \multicolumn{2}{c}{\textbf{WikiTQ}} \\
\cmidrule(lr){2-3} \cmidrule(lr){4-5}
& No Mem & Raw Sib & No Mem & Raw Sib \\
\midrule
Accuracy (\%) & 27.5 & \textbf{39.2} & 26.5 & \textbf{30.6} \\
Apol.\ terminals & 90 & 71 & 64 & 53 \\
Sel.\ apol. & 28 & 23 & 18 & 15 \\
All-apol.\ states & 25 & --- & --- & --- \\
\midrule
$\Delta$ (Raw Sib) & \multicolumn{2}{c}{+11.7\%, $-$21\% apol.} & \multicolumn{2}{c}{+4.1\%, $-$17\% apol.} \\
\bottomrule
\end{tabular}%
}
\end{table}

\paragraph{Mechanism.}
At $T{=}0.7$ sampling, Haiku~3.5 generates apologize-style continuations
in all 3 sibling candidates after encountering SQL errors
(e.g., \texttt{sqlite\_master} syntax error on MySQL).
Greedy decoding ($T{=}0$) avoids this by argmax selecting a retry token
(e.g., ``Let me try \texttt{information\_schema}...''),
but the broader sampling distribution at $T{=}0.7$ collapses
to the apologize mode.
Verified on WikiSQL: 0 mixed states (apologize + retry) exist ---
PRM never faces a retry-vs-apologize choice.

\paragraph{Connection to \citet{chen-etal-2024-tree}.}
Chen et al.\ show that tree search demands discriminator accuracy
$\geq$90\% to improve over re-ranking.
Our finding identifies a complementary failure mode:
even with a perfect discriminator,
tree search cannot help when the \emph{generator} lacks diversity
at critical branching points.
Cross-sibling memory (Raw Sibling) addresses this by conditioning
each candidate on prior siblings' outcomes,
expanding the effective generation distribution beyond
what temperature-based stochastic sampling achieves alone.

%% file: sections/app-kgqa-fact-regression.tex
\section{KGQA Fact Extraction: Regression Case Studies}
\label{app:kgqa-fact-regression}

This appendix examines the per-example regression mechanisms behind
LiTS-Fact's accuracy-neutral effect on KGQA (\S\ref{sec:f3-efficiency}).
The analysis was performed on a 150-example pilot run; the main matrix
in Table~\ref{tab:experiment-matrix} reports KGQA on the 69-example
subset (\S\ref{sec:setup}).
On the 69-subset, KGQA Indep$+$LiTS-Fact rescues 4 examples and
regresses on 4 (Appendix~\ref{app:mcnemar-pvalues}); the case studies
below illustrate the regression mechanisms, four of which (idx 23, 92,
143, 147) fall within the 69-subset.

Of the 150 KGQA examples, LiTS-Fact regresses on 6
(baseline pass@5 $\to$ fact fail).
All 6 share a common mechanism:
facts reduce stochastic diversity by providing a ``free'' path
that the agent follows instead of exploring alternatives.
We identify four specific manifestations on KGQA:

\paragraph{Wrong-type path (idx=23).}
Question: ``What Canadian government exists where Baldur von Schirach was born?''
(Answer: Constitutional monarchy, type \texttt{form\_of\_government}).
Facts record that ``canadian'' has relation \texttt{governmental\_jurisdiction.agencies}
(type \texttt{government\_agency}).
Agent follows this path $\to$ wrong answer type.
Baseline (4/5 pass) independently discovers \texttt{form\_of\_government}.

\paragraph{Procedural instruction (idx=92, 147).}
Question: ``How many games of Electronic Arts are released in the US?'' (Answer: 5).
Facts include a procedural instruction:
``To find games by a publisher in a region, you must: (1) get games\_published,
(2) traverse versions, (3) intersect.''
This indirect path yields 4; the direct relation
\texttt{game\_versions\_published} yields 5.
Baseline (1--2/5 pass) stochastically finds the direct path.

\paragraph{Wrong-relation lock-in (idx=87).}
Question: ``What programming language paradigm is followed by AngelScript
and derivatives of Prolog?'' (Answer: Object-oriented programming, single
entity).
Facts record that ``\texttt{computer.programming\_language.influenced}
on Prolog returns the programming languages that Prolog influenced
(its derivatives/dialects).''
Baseline attempts pick the narrower
\texttt{computer.programming\_language.dialects} relation
(returning only Prolog's dialects) and reach the correct single-entity
intersection. The fact-augmented attempts consistently use the broader
\texttt{influenced} relation, returning a superset whose paradigm
intersection includes one extra non-answer entity, scoring 0 across
all 5 attempts.

\paragraph{Correct-but-irrelevant negative fact (idx=143).}
Question: ``What Facebook founder was influenced by Jeff Bezos?''
(Answer: Mark Zuckerberg).
Facts state: ``Facebook founders do NOT have \texttt{influenced\_by}.''
This is correct (the type lacks that relation),
but the correct path goes from Jeff Bezos $\to$ \texttt{influenced} $\to$ people,
not from founders $\to$ \texttt{influenced\_by}.
Agent over-generalizes the negative fact and avoids the influence direction entirely.

\paragraph{Over-exploration (idx=25).}
Agent finds the correct intermediate answer
but facts record deeper relations explored in attempt~0.
Subsequent attempts follow these ``there's more to explore'' signals
instead of stopping, hitting max\_steps.

%% file: sections/app-conflict.tex
\section{Case Study: Abstraction-Level Conflict}
\label{app:conflict}

This case study illustrates how combining reflection and fact extraction
in the same prompt can degrade both accuracy and efficiency,
using WikiSQL Example~0 (``What are the Notes when the Method is decision?'').

\paragraph{Setup.}
Three configurations are compared, all using Haiku~3.5 as policy
and Sonnet~4.6 as augmentor LLM, with 5 best-of-$N$ at $T{=}0.9$.

\paragraph{Fact extraction alone.}
After attempt~0 (8 steps, calls \texttt{sql\_db\_list\_tables}),
the memory backend extracts environmental facts:
\begin{quote}
\small
\emph{``The database contains a table named `Tournament Results'.''} \\
\emph{``The `Tournament Results' table has a `Notes' column.''} \\
\emph{``The `Tournament Results' table has a `Method' column
that contains values such as `decision'.''} \\
\end{quote}
Attempts~1--4 skip \texttt{sql\_db\_list\_tables} entirely
(the agent already knows which table to query)
and solve the task in 3--4 steps.

\paragraph{Reflection alone.}
After attempt~0 fails, the reflection LLM generates:
\begin{quote}
\small
\emph{``Step~7 omitted the Notes column from the SELECT clause.
Revised Plan: 1.~List all tables. 2.~Check schema.
3.~Run:} \texttt{SELECT Notes FROM <table> WHERE Method = `Decision'}\emph{.
4.~Return the result.''}
\end{quote}
Attempts~1--4 follow this plan and achieve higher accuracy
(pass@5 = 58.8\% across all examples),
but still call \texttt{sql\_db\_list\_tables} in every attempt
because the plan explicitly instructs it.

\paragraph{Fact + Reflection combined.}
Both augmentors inject into the system prompt simultaneously.
The facts say the table is \texttt{Tournament Results};
the reflection says ``1.~List all tables.''
The agent follows the reflection's explicit step-by-step plan,
calling \texttt{sql\_db\_list\_tables} in 4/5 attempts
despite the facts making this step unnecessary.

\begin{table}[ht]
\centering
\small
\caption{Example~0: \texttt{sql\_db\_list\_tables} calls
and step counts across configurations.}
\label{tab:conflict-example}
\begin{tabular}{@{}lccccc@{}}
\toprule
& a0 & a1 & a2 & a3 & a4 \\
\midrule
\multicolumn{6}{@{}l}{\textbf{Fact only} (list\_tables skipped from a1)} \\
Steps       & 8 & 4 & 3 & 3 & 3 \\
list\_tables & \checkmark & --- & --- & --- & --- \\
\midrule
\multicolumn{6}{@{}l}{\textbf{Fact + Reflection} (list\_tables not skipped)} \\
Steps       & 5 & 7 & 7 & 5 & 6 \\
list\_tables & \checkmark & \checkmark & \checkmark & --- & \checkmark \\
\bottomrule
\end{tabular}
\end{table}

\paragraph{Mechanism.}
The conflict arises because the two memory types operate
at different abstraction levels.
Facts provide \emph{implicit} environmental knowledge
(``table X exists with columns A, B, C'')
that the agent can use to skip discovery steps ---
but only if no other signal overrides this.
Reflection provides \emph{explicit} procedural plans
(``Step~1: list tables, Step~2: check schema'')
that the agent follows literally.
When both are present, the explicit plan dominates:
the agent treats the reflection as an instruction sequence
rather than consulting the facts to determine
which steps are already unnecessary.

\paragraph{Aggregate effect.}
Across all 51 WikiSQL examples,
the \texttt{list\_tables} skip rate drops
from 77\% (fact only) to 20\% (fact + reflection)
in attempts~1--4.
Pass@5 drops from 58.8\% (reflection only) to 52.9\%
(fact + reflection),
suggesting that the added noise from conflicting signals
also hurts accuracy.
The same pattern holds on WikiTQ (70\% $\to$ 23\% skip rate;
49.0\% $\to$ 46.9\% pass@5).

%% file: sections/app-retrieval.tex
\section{Retrieval Policies for Cross-Trajectory Fact Memory}
\label{sec:appendix-retrieval}

We expand on the argument in \S\ref{sec:discussion}
that selective retrieval cannot resolve the diversity-vs-efficiency tradeoff.
Table~\ref{tab:retrieval-policies} enumerates four retrieval policies
along the dimensions of diversity preservation
and efficiency (skipping redundant discovery).
Only \emph{Inject All} (LiTS-Fact in our experiments) and \emph{None} (baseline)
are evaluated in this paper;
the other two are included as a design-space analysis,
not empirical results.

\begin{table*}[h]
\centering
\footnotesize
\caption{Retrieval policies for cross-trajectory fact memory.
Only \emph{Inject All} and \emph{None} correspond to evaluated configurations
(LiTS-Fact and No Memory respectively in Table~\ref{tab:experiment-matrix});
\emph{Similar} and \emph{Dissimilar} are predicted behaviors.
\textbf{Diversity}: ability of subsequent attempts to explore alternative paths.
\textbf{Efficiency}: ability to skip redundant environmental discovery.}
\label{tab:retrieval-policies}
\resizebox{\textwidth}{!}{%
\begin{tabular}{@{}l p{3.0cm} p{4.5cm} p{4.5cm}@{}}
\toprule
\textbf{Policy} & \textbf{What is injected} & \textbf{Diversity} & \textbf{Efficiency} \\
\midrule
\emph{Inject All} (LiTS-Fact)
  & Union of all prior trajectories' facts
  & Reduced --- agent anchors on the most-supported prior path
  & High --- known facts let the agent skip redundant discovery \\
\addlinespace
\emph{Similar to current state} (predicted)
  & Facts most embedding-similar to the current step
  & Further reduced --- retrieval surfaces precisely the prior path
  & Inconsistent --- relevant fact retrieved only when current step is surface-similar \\
\addlinespace
\emph{Dissimilar to current state} (predicted)
  & Facts least similar to the current step
  & Could push exploration; could also distract
  & Low --- facts that enable shortcuts are systematically excluded \\
\addlinespace
\emph{None} (No Memory baseline)
  & ---
  & Maximal --- attempts are i.i.d.\ samples from the policy
  & Low --- every attempt re-discovers environmental structure \\
\bottomrule
\end{tabular}%
}
\end{table*}

\paragraph{Why no policy resolves the tradeoff.}
Diversity loss is triggered by the \emph{presence} of any positive
prior facts in the prompt, not by which subset is chosen ---
once injected, the LLM treats them as ground truth and biases toward
consistency.
Conversely, the efficiency benefit of fact memory comes from
making environmental knowledge available across trajectories ---
which similarity-based retrieval undermines whenever the relevant
fact's discovery context differs from the current step's context.
The Pareto frontier in this design space is therefore narrow,
and our \emph{Inject All} configuration sits on the high-efficiency,
low-diversity end of it.
Mitigations identified in \S\ref{sec:discussion}
(negative-fact extraction, candidate-vs-truth framing)
operate on \emph{what is stored} or \emph{how it is presented},
not on the retrieval axis.

%% file: sections/app-efficiency.tex
\section{Efficiency Metrics}
\label{app:efficiency}

\begin{table}[h]
\centering
\small
\caption{Efficiency metrics for best-of-$N$.
\emph{Steps}: mean trajectory length (lower is better).
\emph{Skip}: fraction of attempts 1--4 that skip the initial
discovery call (higher = more efficient).
For SQL: skipping \texttt{list\_tables}; for KG: not applicable.
\emph{Cost}: policy cost / total cost per 5-attempt run.
Total includes augmentor LLM (Sonnet \$3/\$15).
SQL policy: Haiku \$0.80/\$4.00; KG policy: Sonnet \$3/\$15.}
\label{tab:efficiency}
\begin{tabular}{@{}l l cc c@{}}
\toprule
& & \textbf{Steps} & \textbf{Skip \%} & \textbf{Cost (pol.\ / tot.)} \\
\midrule
\multirow{4}{*}{\rotatebox{90}{\small WikiSQL}}
& No Memory       & 6.1 &  4 & 2.20 / 2.20 \\
& Reflection      & 6.4 &  5 & 3.26 / 5.02 \\
& LiTS-Fact       & \textbf{4.9} & \textbf{77} & 1.68 / 2.66 \\
& Fact + Refl.    & 5.9 & 20 & 3.54 / 5.89 \\
\midrule
\multirow{4}{*}{\rotatebox{90}{\small WikiTQ}}
& No Memory       & 5.8 &  5 & 1.89 / 1.89 \\
& Reflection      & 5.9 &  8 & 2.71 / 4.35 \\
& LiTS-Fact       & \textbf{4.7} & \textbf{70} & 1.50 / 2.34 \\
& Fact + Refl.    & 5.6 & 23 & 2.75 / 5.28 \\
\midrule
\multirow{3}{*}{\rotatebox{90}{\small KGQA}}
& No Memory       & 8.9 & --- & 60 / 60 \\
& Reflection      & 8.9 & --- & 76 / 82 \\
& LiTS-Fact       & \textbf{6.6} & --- & 58 / 64 \\
\bottomrule
\end{tabular}
\end{table}

%% file: sections/app-compute-cost.tex
\section{Compute Cost}
\label{app:compute-cost}

Table~\ref{tab:compute-cost} reports the API cost and token usage of
each configuration.
All costs are computed from logged token counts at Bedrock pricing:
Haiku~3.5 (\$0.80/\$4.00 per 1M input/output tokens) for
WikiSQL/WikiTQ/Terminal-Bench policy; Sonnet~4.6 (\$3.00/\$15.00)
for KGQA policy, all per-step reward models, and all
memory/augmentor LLMs.
Raw Sibling does not invoke a separate LLM (it injects raw
observations directly), so its cost equals the No-Memory baseline.
The total experiment cost is approximately \textbf{\$1,384}.

\begin{table}[h]
\centering
\footnotesize
\caption{Per-configuration API cost and token usage.
$N$ = evaluated examples.
\emph{Pol.} = policy LLM calls/tokens;
\emph{Sup.} = supervisor (PRM + memory/augmentor) calls/tokens.
KGQA Indep runs on 150 examples; Beam/MCTS on 69-subset.
\textsuperscript{\dag}Not evaluated (see
Table~\ref{tab:experiment-matrix}).}
\label{tab:compute-cost}
\resizebox{\columnwidth}{!}{%
\begin{tabular}{@{}ll r rr rr c@{}}
\toprule
& & & \multicolumn{2}{c}{\textbf{Policy}} & \multicolumn{2}{c}{\textbf{Supervisor}} & \\
\cmidrule(lr){4-5} \cmidrule(lr){6-7}
\textbf{Bench.} & \textbf{Config} & $N$ & calls & tok (M) & calls & tok (M) & \textbf{Cost} \\
\midrule
\multirow{9}{*}{\rotatebox{90}{\scriptsize WikiSQL}}
& Indep No-Mem     & 51  & 1310 & 2.2 & --- & --- & \$2 \\
& Indep Refl       & 51  & 1379 & 3.5 & 255 & 0.2 & \$5 \\
& Indep Fact       & 51  & 985 & 1.6 & 255 & 0.2 & \$3 \\
& Beam No-Mem      & 51  & 723 & 1.2 & 714 & 1.1 & \$7 \\
& Beam Raw Sib     & 51  & 729 & 1.2 & 724 & 1.1 & \$7 \\
& MCTS No-Mem      & 51  & 2880 & 5.4 & 2879 & 5.1 & \$31 \\
& MCTS Refl        & 51  & 2910 & 7.3 & 2910 & 5.1 & \$31 \\
& MCTS Raw Sib     & 51  & 2871 & 5.4 & 2870 & 5.1 & \$31 \\
& MCTS Fact        & 51  & 3027 & 6.1 & 6054 & 6.4 & \$40 \\
\midrule
\multirow{6}{*}{\rotatebox{90}{\scriptsize WikiTQ}}
& Indep No-Mem     & 49  & 1183 & 1.9 & --- & --- & \$2 \\
& Indep Refl       & 49  & 1197 & 2.9 & 245 & 0.2 & \$4 \\
& Indep Fact       & 49  & 909 & 1.5 & 245 & 0.1 & \$2 \\
& MCTS No-Mem      & 49  & 2619 & 4.8 & 2619 & 4.5 & \$27 \\
& MCTS Refl        & 49  & 2682 & 6.7 & 2900 & 4.8 & \$31 \\
& MCTS Raw Sib     & 49  & 2454 & 4.5 & 2454 & 4.2 & \$26 \\
\midrule
\multirow{8}{*}{\rotatebox{90}{\scriptsize KGQA}}
& Indep No-Mem     & 150 & 6698 & 16.9 & --- & --- & \$60 \\
& Indep Refl       & 150 & 6234 & 21.5 & 750 & 0.7 & \$82 \\
& Indep Fact       & 150 & 4947 & 15.6 & 750 & 1.5 & \$64 \\
& Beam No-Mem      & 69  & 2202 & 6.0 & 2202 & 3.9 & \$39 \\
& Beam Raw Sib     & 69  & 1794 & 5.3 & 1794 & 3.4 & \$36 \\
& MCTS No-Mem      & 69  & 9043 & 26.9 & 9042 & 18.9 & \$186 \\
& MCTS Refl        & 69  & 7953 & 28.9 & 8298 & 16.5 & \$183 \\
& MCTS Raw Sib     & 69  & 8163 & 26.9 & 8163 & 18.0 & \$188 \\
\midrule
\multirow{3}{*}{\rotatebox{90}{\scriptsize T-B}}
& Indep No-Mem     & 89  & 6675 & 96.4 & --- & --- & \$85 \\
& Indep Refl       & 89  & 6952 & 117.8 & 369 & 2.4 & \$112 \\
& Indep Fact       & 89  & 6461 & 110.4 & 440 & 2.4 & \$105 \\
\bottomrule
\end{tabular}%
}
\end{table}